\documentclass[11pt,a4paper]{article}
\usepackage[hyperref]{naaclhlt2019}
\usepackage{times}
\usepackage{latexsym}
\usepackage{url}
\usepackage{boldline}

\usepackage{amsmath}
\usepackage{multirow}
\usepackage{graphicx}
\newcolumntype{C}[1]{>{\centering\arraybackslash}m{#1}}

\aclfinalcopy % Uncomment this line for the final submission
\title{Expansional Retrofitting for Word Vector Enrichment}

\author{Hwiyeol Jo\\
  AI Lab, LG Electronics \\
  {\tt hwiyeolj@gmail.com}}
\date{}

\begin{document}
\maketitle
\begin{abstract}
    Retrofitting techniques, which inject external resources into word representations, have compensated the weakness of distributed representations in semantic and relational knowledge between words. Implicitly retrofitting word vectors by expansional technique outperforms retrofitting in word similarity tasks with word vector generalization. In this paper, we propose {\em unsupervised extrofitting}: expansional retrofitting (extrofitting) without external semantic lexicons. We also propose {\em deep extrofitting}: in-depth stacking of extrofitting and further combinations of extrofitting with retrofitting. When experimenting with GloVe, we show that our methods outperform the previous methods on most of word similarity tasks while requiring only synonyms as an external resource. Lastly, we show the effect of word vector enrichment on text classification task, as a downstream task.
\end{abstract}

\section{Introduction}
\subsection{Motivation}
    Distributed word representation is widely used to compute the similarity of words and word relations (e.g., mean square distance, cosine similarity). Most of the algorithms to generate distributed representation are based on the basic idea of CBoW (Continuous Bag-of-Words) and skip-gram~\cite{mikolov2013distributed}. Both algorithms learn word vectors by maximizing the probability of occurrence of a center word given neighbor words or neighbor words given a center word.\\
    Due to the aforementioned nature, the distributed word representation is weak at representing semantic and relational meanings of words that cannot be captured by the word orders~\cite{lenci2018distributional}. In order to inject semantic information, 2 types of approaches were suggested: heavy-weight approach and post-processing approach.\\
    The heavy-weight approach is to modify the objective function of word embedding algorithms to reflect semantic information when generating word vector from raw text. However, the heavy-weight approach is less competitive because of relatively small improvement in performance, compared to computational complexity.\\
    The post-processing method, called retrofitting, is to inject the semantic information of external resources by modifying the values of pretrained word vectors. The benefits of post-processing method are that (1) it can reflect additional resources into the word vectors without re-training on all the data, (2) retrofitting can be applied to all kinds of pretrained word vectors, and (3) retrofitting can modify word vectors to specialize in a specific task. For example, when retrofitting is applied to sentiment analysis on movie domain, it aggregates least relevant word vectors like movie titles, characters, and other entities that the sentiment analysis model can be more dependent on sentiment words such as {\tt like}, {\tt favorite}.\\
    However, to enrich word vectors with post-processing methods, it is necessary to define semantic lexicons. There are publicly opened semantic lexicons but old-fashioned or having only a few words. Even the semantic lexicons are difficult to update because we need expert knowledge on word meanings.\\
    In this work, we first suggest {\em unsupervised extrofitting}, which enriches word vectors using semantically related words extracted from themselves instead of external semantic lexicons. The method uses Latent Semantic Analysis (LSA)~\cite{landauer1997solution} to extract semantically related words and then applies expansional retrofitting (extrofitting)~\cite{jo2018extrofitting} the word vectors with the information. Unsupervised extrofitting performs on par with extrofitting, which requires external semantic lexicon. Next, we propose {\em deep extrofitting}, which is in-depth stacking of extrofitting and further combination of extrofitting and retrofitting for word vector specialization. The methods prevent retrofitting from converging in a few iterations by extrofitting, finding new enriched vector space. Deep extrofitting outperforms previous methods on word vector specialization while requiring only synonyms.
    
\subsection{Previous Works}
    The first successful post-processing approach is \citeauthor{faruqui2015retrofitting}'s retrofitting, which modifies word vectors by weighted averaging the word vectors with semantic lexicons. They extracted synonym pairs from PPDB~\cite{ganitkevitch2013ppdb}, WordNet~\cite{miller1995wordnet}, and FrameNet~\cite{baker1998berkeley}, and use them as external resources. The retrofitting dramatically improves word similarity between synonyms, and the result not only corresponds to human intuition on words but also performs better on document classification tasks compared to the original word embeddings~\cite{kiela2015specializing}. After that, \citeauthor{mrkvsic2016counter} proposed counter-fitting, which uses synonym pairs to collect word vectors and antonym pairs to make word vectors distant from one another. The counter-fitting showed good performance at word vector specialization. Next, ATTRACT-REPEL~\cite{mrkvsic2017semantic} was suggested that the model injects linguistic constraints into word vectors by learning from predefined cost function with mono- and cross-lingual synonym and antonym constraints. Explicit Retrofitting~\cite{glavavs2018explicit} directly learns mapping functions of linguistic constraints with deep neural network architecture and retrofits the word vectors.\\
    The previous researches focused on explicit retrofitting, using manually defined or learned function to make synonyms close or antonyms distant. As a result, their approaches were strongly dependent on external resources and pretrained word vectors. Furthermore, we believe that making synonyms close together is reasonable even though it has different nuance in some context, but antonyms have to be further investigated rather than making them afar. For example, {\tt love} and {\tt hate} are grouped as antonyms, but they should share the meaning of `emotion' in their representation. Lastly, the usefulness of word vector specialization should also be further investigated. Previous works showed that specialized word vectors improve the performance of domain-specific downstream tasks, but they did not show the effect of word vector post-processing on conventional NLP tasks such as text classification.\\
    \citeauthor{jo2018extrofitting} presented extrofitting, a method to enrich not only word representation but also its vector space using semantic lexicons. The method overcomes dependency problems on pretrained word vector and explicit functions in that extrofitting implicitly retrofits word vectors by expanding and reducing its dimensions, without explicit retrofitting functions. While adjusting the dimension of vector space, the algorithm could strengthen the meaning of each word, making synonyms close together and non-synonyms far from each other, finally projecting the new vector space in accordance to the distribution of word vectors. Therefore, extrofitting resolves the issue of using antonyms and explicit retrofitting function.
    
\subsection{Contributions}

    Our main contributions can be summarized as follows:
    \begin{itemize}
        \item We propose {\em unsupervised extrofitting} that extends extrofitting for enriching word vectors without using external semantic lexicons. The method can resolve one of the limitations of post-processing approaches, which requires well-defined semantic lexicon.
        \item We also propose {\em deep extrofitting} that extends extrofitting for word vector specialization. This simple extension outperforms previous methods while requiring only synonyms. 
        \item We report the effects of word vector post-processing on conventional text classification task. The result shows that our methods are `enrichment', which improves the performance on conventional tasks when compared to word vector specialization methods.
    \end{itemize}
    
\section{Preliminary}

\subsection{Retrofitting}
    Retrofitting~\cite{faruqui2015retrofitting} defines an objective function $\Psi(Q)$ that make synonym pairs in semantic lexicon close together. The algorithm learns the retrofitted word embedding matrix $Q = \{q_1 , q_2 , \dots , q_n\}$ as follows:
    \begin{equation*}
    \Psi (Q) = \sum_{i=1}^n \ [\alpha || q_i - \hat{q}_i ||^2 + \sum_{(i,j) \in E} \beta_{i j} ||q_i - q_j||^2]
    \end{equation*}
    where an original word vector is ${q}_i$, its synonym vector is $q_j$, inferred word vector is $\hat{q}_i$, and $E$ denotes synonym pairs in semantic lexicons. The hyperparameter $\alpha$ and $\beta$ control the relative strengths of associations.

\subsection{Latent Semantic Analysis (LSA)}
    LSA~\cite{landauer1997solution} has been used to extract the relation of data through latent variables. LSA is based on Singular Value Decomposition (SVD), which decomposes a matrix as follows:
    \begin{equation*}
        A = U S V^T,
    \end{equation*}
    where $S$ is  a diagonal matrix with singular values, and $U$ and $V$ are the orthogonal eigenvectors. We can select top-$k$ singular values to represent matrix $A$ in $k$-dimensional latent space. Then $U$ and $V$ are redefined as $U_k \in \mathcal{R}^{N \times k}$ and $V_k \in \mathcal{R}^{k \times N}$, respectively, with diagonal matrix $S_k \in \mathcal{R}^{k \times k}$.
    When applying LSA in topic modeling, $A$ is defined as term-document matrix. Then, $U S$ and $S V^T$ are considered as term vectors and document vectors in latent space, respectively.

\subsection{Expansional Retrofitting (Extrofitting)}
    Extrofitting first expands word embedding matrix $W$:
    % \begin{equation}
    \begin{multline*}
        \textrm{Expand}(W) \\ = W \oplus r_w 
            \left\{ 
            \begin{array}{cl}
                \textrm{mean}_{w \in syn} (\mu_{w}) & \textrm{if}\ w \in \textrm{L} \\
                \mu_w & \textrm{otherwise}
            \end{array}\right.
    \end{multline*}
    % \end{equation}
    where $\mu_w$ is the mean value of elements in word vector $w$. L denotes semantic lexicons, and $syn$ denotes synonym pairs. Next, we define \textrm{Trans}$(W)$ as calculating transform matrix given word embedding matrix $W$:
    \begin{multline*}
        \textrm{Trans}(W) \\ = \textrm{argmax}_U \frac{| U^T \sum_{c} (\mu_c - \mu) (\mu_c - \mu)^T U |}{| U^T \sum_{c} \sum_{i} (x_i - \mu_c) (x_i - \mu_c)^T U |}
    \end{multline*}
    where $x$ is a word vector, $c$ is a class. The overall average of $x$ is $\mu$, and the class average in class $i$ is denoted by $\mu_i$. This formula finds transform matrix $U$ which minimizes the variance within the same class and maximizes the variance between different classes. Each class is defined as the index of synonym pairs. Then simple extrofitting is formulated as follows:
    \begin{equation*}
        \textrm{Extro}(W) = \textrm{Trans}(\textrm{Expand}(W))^T \textrm{Expand}(W)
    \end{equation*}
    
\section{Unsupervised Extrofitting}
    We consider word embedding matrix as the term-document matrix of LSA. Specifically, the word embedding matrix is term-semantic matrix per se so we expect to extract semantically related words (terms) by matrix decomposition. We first decompose word embeddings $W$ as follows:
    \begin{equation*}
        W_k = U_k S_k V_k^T
    \end{equation*}
    We can get word representations in latent space ($W'$) by computing $U_k S_k$. By comparing each word in latent space, we extract semantically related words using cosine similarity. Then, we define the set of semantically related words as the class $c$ of extrofitting:
    \begin{equation*}
        \begin{array}{ll}
            c_{w_i} = c_{w_j}, & \textrm{if}\ cos( W'_i , W'_j ) \ge T \\
        \end{array}
    \end{equation*}
    where $w$ is a word, and $T$ is a threshold that determines the words $w_i$ and $w_j$ are semantically related. In our experiment, we set the threshold to high value (0.9) since type II error is rather better than type I error.
    
\section{Deep Extrofitting with Semantic Lexicon}
    
\subsection{Stacked Extrofitting}
    We first generalize extrofitting. The stacked extrofitting (Extro$_{Iter}(W)$) is formulated as follows:
    \begin{multline*}
        \textrm{Extro}_n(W) \\= \textrm{Trans}(\textrm{Extro}_{n-1}(W))^T \textrm{Expand}(\textrm{Extro}_{n-1}(W)).
    \end{multline*}
    % Next, we do not keep the original dimension. That is, we skip Step 1 (expanding word vector with enrichment) and Step 2 (transferring semantic knowledge), which originally keep the original dimension of word vectors. Thereby we can focus on the effect of Step 3, enriching vector space by reducing its vector dimension. The stacked extrofitting without keeping dimension is formulated as follows:
    % \begin{multline*}
    %     \textrm{Extro}_n(W) \\= \textrm{Trans}(\text{Extro}_{n-1}(W))^T \textrm{Extro}_{n-1}(W).
    % \end{multline*}
    
\subsection{Extrofitting with Retrofitting}
    
    Retrofitting could be specialized in semantic lexicons whereas extrofitting results in generalized word vectors~\cite{jo2018extrofitting}. If then, we expect the results of retrofitting and extrofitting to complement each other. So, we apply retrofitting to word vectors and then extrofit the retrofitted word vectors, and vice versa. We denote retrofitting as Retro$_{Iter}(W)$.
    
    \begin{equation}
        \begin{aligned}
        \text{RExtro}_{nm}(W) &= \text{Extro}_m(\text{Retro}_n(W))\\
        \text{ERetro}_{nm}(W) &= \text{Retro}_m(\text{Extro}_n(W))
        \notag
        \end{aligned}
    \end{equation}
    \noindent Further, we can use them one by one:
    \begin{equation}
        \begin{aligned}
        \text{Stepwise RExtro}_n = \{\text{Extro}_1(\text{Retro}_1(W))\}_n \\
        \text{Stepwise ERetro}_n = \{\text{Retro}_1(\text{Extro}_1(W))\}_n
        \notag
        \end{aligned}
    \end{equation}

\section{Experiment Data}
\subsection{Pretrained Word Vector}
    Pretrained word vectors include words composed of n-dimensional float vectors. One of major pretrained word vector we used is {\bf GloVe}~\cite{pennington2014glove}. We use {\tt glove.42B.300d} trained on Common Crawl data, which contains 1,917,493 unique words as 300 dimensional vectors.\\
    Even though many word embedding algorithms and pretrained word vectors have been suggested after GloVe, GloVe is still being used as a strong baseline on word similarity tasks~\cite{cer2017semeval,camacho2017semeval}. We also use {\bf Word2Vec}~\cite{mikolov2013efficient}, {\bf Fasttext}~\cite{bojanowski2016enriching}, and {\bf Paragram}~\cite{wieting2015paraphrase} as resources of unsupervised extrofitting.
    
\subsection{Semantic Lexicon}
    As an external semantic lexicon, we use {\bf WordNet}~\cite{miller1995wordnet}, which consists of approximately 150,000 words and 115,000 synsets pairs. We borrow \citeauthor{faruqui2015retrofitting}'s WordNet$_{all}$ lexicon, comprised of synonyms, hypernyms, and hyponyms. 
    % WordNet$_{all}$ overlaps 70,411 words with GloVe, which is 3.67\% of words in GloVe.
    \citeauthor{faruqui2015retrofitting} reported that their method performed the best when paired with WordNet$_{all}$. Extrofitting~\cite{jo2018extrofitting} also worked well with WordNet$_{all}$. 

\subsection{Intrinsic Evaluation Dataset}
    Word similarity datasets consist of two word pairs with human-rated similarity score between the words. With the datasets, word similarity task is defined as calculating Spearman's correlation~\cite{daniel1990spearman} between two words in word vector format. We use 4 different kinds of datasets: {\bf MEN-3k (MEN)}~\cite{bruni2014multimodal}, {\bf WordSim-353 (WS)}~\cite{finkelstein2001placing}, {\bf SimLex-999 (SL)}~\cite{hill2015simlex}, and {\bf SimVerb-3500 (SV)}~\cite{gerz2016simverb}.\\We experiment our methods on as many datasets as possible to see the effect of word vector enrichment while avoiding to become overfitted to a specific dataset. When we use MEN-3k, WordSim-353, and SimVerb-3500, we combine train (or dev) set and test set together solely for evaluation. The other datasets are left for future work since the datasets either are too small or contain numerous out-of-vocabulary words.
    
\section{Experiments}

\subsection{Unsupervised Extrofitting}
    In Table~\ref{tab:1}, our method improves the performance on all the word similarity dataset when compared to GloVe. The result implies that pretrained word vector can be enriched by unsupervised extrofitting, which does not require any semantic lexicon. Even unsupervised extrofitting outperforms on MEN when compared to extrofitted GloVe with external semantic resource (WordNet).\\
    We also experiment that the extracted semantic information can be used to retrofitting but the information is not useful for retrofitting, which means the extracted information cannot be considered as synonyms (see Appendix~\ref{appendix:a}). Therefore, we consider them as `semantically related' words.
    
    \begin{table}[ht!]
    \begin{center}
    \begin{tabular}{|l||c|c|c|c|} \hlineB{3}
    & \bf MEN & \bf WS & \bf SL & \bf SV \\ \hlineB{3}
    % \multicolumn{5}{|l|}{\tt Raw} \\ \hline
    GloVe & .7435 & .5516 & .3738 & .2264 \\ \hlineB{3}
    % \multicolumn{5}{|l|}{\tt Unsupervised Retrofitting} \\ \hline
    % +GloVe(50) & . & . & . & . \\ \hline
    % +GloVe(100) & .7435 & .5516 & .3738 & .2245 \\ \hline
    % +GloVe(150) & . & . & . & . \\ \hline
    % +GloVe(200) & .7435 & .5516 & .3738 & .2247 \\ \hline
    % +GloVe(300) & .7435 & .5516 & .3738 & .2257 \\ \hlineB{3}
    \multicolumn{5}{|l|}{\tt Unsupervised Extrofitting} \\ \hline
    % +GloVe(50) &  .8212 & .6319 & \bf .4933 & .3341 \\ \hline
    +GloVe(50) &  .8084 & .6010 & \bf .4775 & .3077 \\ \hline
    +GloVe(100) & \bf .8271 & \bf .6506 & .4754 & \bf .3382 \\ \hline
    % +GloVe(150) & .8135 & .6421 & .4589 & \bf .3250 \\ \hline
    +GloVe(150) & .8033 & .6223 & .4459 & .2980 \\ \hline
    +GloVe(200) & .7939 & .6091 & .4287 & .2818 \\ \hline
    +GloVe(300) & .7900 & .6037 & .4439 & .2936 \\ \hline
    \hlineB{3}
    \multicolumn{5}{|l|}{\tt Extrofitting with Lexicons} \\ \hline
    +WordNet & .8215 & \bf .6552 & \bf .4930 & \bf .3596 \\ \hline
    % \hlineB{3}
    % Fasttext & \bf .7654 & \bf .6304 & .3803 & .2581 \\ \hline
    % +Fasttext(300) & .7469 & .6285 & \bf .4020 & .2839 \\ \hline
    % +Fasttext(200) & .7460 & .6281 & .4019 & \bf .2852 \\ \hline
    % +Fasttext(150) & .7453 & \bf .6304 & .4008 & .2838 \\ \hline
    % +Fasttext(100) & .7442 & .6178 & .3989 & .2817 \\ \hline
    % +Fasttext(50) & .7421 & .6269 & .3952 & .2786 \\ \hline
    % \hlineB{3}
    % Word2Vec & \bf .7764 & \bf .6156 & .4475 & . \\ \hline
    % +Word2Vec(300) & .4795 & .2957 & .2005 & .1063 \\ \hline
    % +Word2Vec(200) & . & . & . & \bf . \\ \hline
    % +Word2Vec(150) & .7453 & \bf .6304 & .4008 & .2838 \\ \hline
    % +Word2Vec(100) & . & . & . & . \\ \hline
    % +Word2Vec(50) & .7421 & .6269 & .3952 & .2786 \\ \hline
    \end{tabular}
    \end{center}
    \caption{Spearman's correlation of unsupervised extrofitted GloVe. We combine train (or dev) set and test set of word similarity dataset together solely for evaluation. GloVe(N) denotes extracted semantic information from GloVe in N-dimensional latent space.}
    \label{tab:1}
    \end{table}
    
    \begin{table*}[ht!]
    \begin{center}
    \begin{tabular}{|c|l||c|c|c|c|}
    \hlineB{3} & & \bf MEN & \bf WS & \bf SL & \bf SV \\ \hlineB{3}
    & GloVe & .7435 & .5516 & .3738 & .2264 \\ \hlineB{3}
    % \multicolumn{5}{|l|}{\tt Unsupervised} \\ \hline
    % +Word2Vec(100) & . & . & . & . \\ \hline
    % +Word2Vec(200) & . & . & . & . \\ \hline
    \multirow{4}{*}{{\small Single}} & +Word2Vec(300) & .6434 & .4921 & .3048 & .1903 \\ \cline{2-6}
    % +Fasttext(100) & . & . & . & . \\ \hline
    % +Fasttext(200) & . & . & . & . \\ \hline
    & +Fasttext(300) & .7749 & .6000 & .4207 & .2698 \\ \cline{2-6}
    % +Fasttext(300)+GloVe(300) & .7907 & .6011 & .4393 & .2924 \\ \hline
    % +GloVe(300) & .8064 & .6333 & .4471 & .3163 \\ \hline
    % +GloVe(300)+Fasttext(300) & .7749 & .5992 & .4205 & .2685 \\ \hline
    % +PgWS(100) & .8280 & .6622 & .4582 & .3399 \\ \hline
    % +PgWS(200) & .8304 & .6699 & .4664 & .3473 \\ \hline
    & +PgWS(300) & \bf .8358 & \bf .6804 & \bf .4685 & \bf .3526 \\ \cline{2-6}
    % +PgSL(100) & .8166 & .6434 & .4409 & .3135 \\ \hline
    % +PgSL(200) & .8200 & .6533 & .4455 & .3184 \\ \hline
    & +PgSL(300) & .8277 & .6669 & .4526 & .3310 \\ \cline{2-6}
    \hlineB{3}
    \multirow{9}{*}{{\small Multi}} 
    & +Fasttext(300)+PgWS(300) & .8365 & .6792 & .4724 & .3578 \\ \cline{2-6}
    & +Fasttext(300)+PgSL(300) & .8304 & .6682 & .4578 & .3375 \\ \cline{2-6}
    & +PgWS(300)+PgSL(300) & .8285 & .6676 & .4539 & .3316 \\ \cline{2-6}
    & +PgSL(300)+PgWS(300) & \bf .8369 & \bf .6884 & \bf .4754 & \bf .3662 \\ \cline{2-6}
    & +GloVe(100)+PgWS(300) & .8357 & .6829 & .4699 & .3538 \\ \cline{2-6}
    & +GloVe(100)+PgSL(300) & .8285 & .6707 & .4539 & .3316 \\ \cline{2-6}
    & +GloVe(100)+PgWS(300)+PgSL(300) & .8345 & .6753 & .4646 & .3508 \\ \cline{2-6}
    & +GloVe(100)+PgSL(300)+PgWS(300) & .8359 & .6786 & .4697 & .3530 \\ \cline{2-6}
    % & +GloVe(100)+PgSL(300)+PgWS(300)+WN & .8221 & .6599 & .4862 & .3571 \\ \cline{2-6}
    % & +PgSL(300)+PgWS(300)+GloVe(300) & .8150 & .6509 & .4610 & .3345 \\ \cline{2-6}
    % & +PgSL(300)+PgWS(300)\_Merged & .8353 & .6808 & .4706 & .3543 \\ \cline{2-6}
    
    % & +GloVe(100)+PgWS(300)+PgWS(300) & \bf .8386 & \bf .6809 & \bf .4770 & \bf .3670 \\ \cline{2-6}
    \hlineB{3}

    % \multicolumn{5}{|l|}{\tt Supervised} \\ \hline
     \multirow{1}{*}{{\small Supervised}} & +WordNet & .8215 & .6552 & \bf .4930 & .3596 \\ \cline{2-6}
    % & +PgWS(300)+WordNet & .8221 & .6669 & .4858 & .3575 \\ \cline{2-6}
    % & +GloVe(100)+PgWS(300)+PargramSL(300)+WN & .8271 & .6799 & .4962 & .3765 \\ \cline{2-6}
    \hlineB{3}
    \end{tabular}
    \end{center}
    \caption{The performance of unsupervised extrofitting ensemble. {\tt PretrainedResource}(N) denotes extracted semantic information from {\tt PretrainedResource}. ParagramWS and ParagramSL are denoted as PgWS and PgSL, respectively.} \label{tab:2}
    \end{table*}
    
    \noindent After observing that our unsupervised extrofitting works well, we borrow other well-known pretrained word vectors in order to use them as semantic resources for unsupervised extrofitting. The results of unsupervised extrofitting ensemble are presented in Table~\ref{tab:2}. Unsupervised extrofitting can utilize other pretrained word vector resources, performing even better than extrofitting with external semantic lexicons.

\subsection{Deep Extrofitting}

    \begin{figure}[ht] \centering
    \includegraphics[scale=0.43]{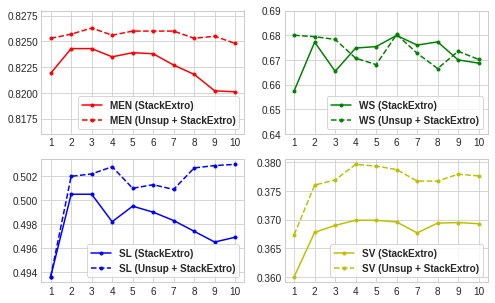}
    \caption{The performance of stacked extrofitting on word similarity tasks. The x-axis indicates iterations.
    % Upper left is MEN, upper right is WS, lower left is SL, and lower right is SV.
    The dotted lines denote the performance of stacked extrofitting after unsupervised extrofitting.}
    \label{fig:1}
    \end{figure}
    
    \begin{figure}[ht] \centering
    \includegraphics[scale=0.43]{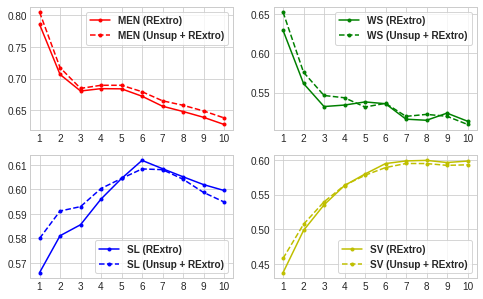}
    \caption{The performance of Stepwise RExtro on word similarity tasks. The x-axis indicates iterations.
    % Upper left is MEN, upper right is WS, lower left is SL, and lower right is SV.
    The dotted lines denote the performance of Stepwise RExtro after unsupervised extrofitting.}
    \label{fig:3}
    \end{figure}
    
    \begin{figure*}[ht] \centering
    \includegraphics[scale=0.39]{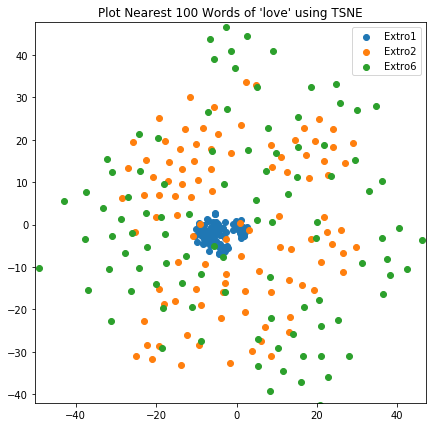}
    \includegraphics[scale=0.39]{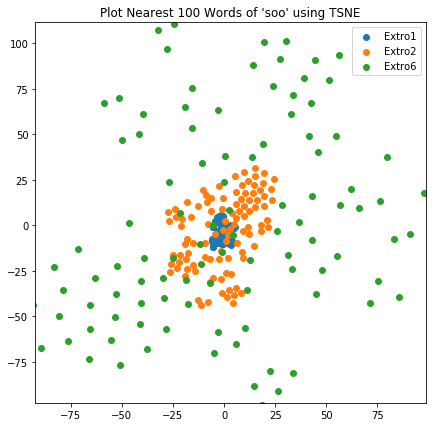}
    \caption{Plots of nearest top-100 words of cue words in stacked extrofitting. We choose two cue words; one is included in semantic lexicons ({\tt love}; left), and another is not ({\tt soo}; right)}
    \label{fig:2}
    \end{figure*}
    
    \begin{figure}[ht] \centering
    \includegraphics[scale=0.43]{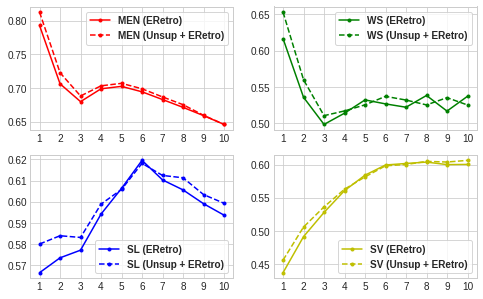}
    \caption{The performance of Stepwise ERetro on word similarity tasks. The x-axis indicates iterations.
    % Upper left is MEN, upper right is WS, lower left is SL, and lower right is SV.
    The dotted lines denote the performance of Stepwise ERetro after unsupervised extrofitting.}
    \label{fig:4}
    \end{figure}
    
    We present the performance of stacked extrofitting (Extro$_n$) to GloVe in Figure~\ref{fig:1}. While stacked extrofitting improves the performance of word similarity tasks for a few iterations, the performance gap becomes smaller as we stack more extrofitting.
    \noindent We also plot top-100 nearest words using t-SNE~\cite{maaten2008visualizing}, as shown in Figure~\ref{fig:2}. Stacking more extrofitting makes the word vectors utilize broader vector space in general while relatively collecting synonyms together. As a result, we might lose word similarity score (see Appendix~\ref{appendix:b}) but gain overall performance improvement. We interpret the results as generalization in that the word vectors get generalized representation by being far away from each other.\\
    % The results show that adding retrofitting more than once does not significantly improves the performance. Also, stacking more extrofitting improves the performance, but the performance gap becomes smaller as we add more extrofitting. We also observe that using 1 retrofitted word vector shows the best performance (see Appendix~\ref{sec:appendix}).\\
    In order to extend our method to word vector specialization, we stack retrofitting and extrofitting, one by one. When we stack retrofitting first, we denote it as Stepwise RExtro$_n$. Otherwise, stacking extrofitting first, we denote it as Stepwise ERetro$_n$. We report the results in Figure~\ref{fig:3} and Figure~\ref{fig:4}, respectively. Stepwise RExtro and Stepwise ERetro perform word vector specialization on SimLex-999 and SimVerb-3500 datasets. Since word pairs in the datasets 100\% overlaps with synonym pairs in WordNet$_{all}$, applying retrofitting improves the similarity on those datasets while concurrently degrading the performance on the other datasets. The performance of retrofitting converges in a few iterations~\cite{faruqui2015retrofitting} but we can specialize over retrofitting with the help of extrofitting by finding new enriched vector space at every iteration, as shown in Table~\ref{tab:3}. Moreover, the weakness of extrofitting--not being able to strongly collect word vectors--is compensated by retrofitting.\\
    We also apply deep extrofitting to the enriched word vectors by unsupervised extrofitting. The results are presented as dotted lines in Figure~\ref{fig:1}, Figure~\ref{fig:3}, and Figure~\ref{fig:4}, showing slightly better performance than original deep extrofitting.
    
    \begin{table*}[ht] \centering
    \begin{tabular}{|c||c|c|}
    \hlineB{3}
        {\small Word} & {\small Method} & Top-10 Nearest Words(Cosine Similarity Score) \\
    \hlineB{3}
    \multirow{7}{*}{{\small\tt love}}   & Raw &
        \begin{tabular}{@{}c@{}c@{}}
             {\small loved(.7745), i(.7338), loves(.7311), know(.7286), loving(.7263),}\\
             {\small really(.7196), always(.7193), want(.7192), hope(.7127), think(.7110)}
        \end{tabular} \\
        \cline{2-3}
                            & {+ Retro} &
        \begin{tabular}{@{}c@{}c@{}}
             {\small loved(.7857), know(.7826), like(.7781), want(.7736), i(.7707),}\\
             {\small feel(.7550), wish(.7549), think(.7491), enjoy(.7453), loving(.7451)}
        \end{tabular} \\
        \cline{2-3}
                            & {\small +SRExtro$_6$} &
        \begin{tabular}{@{}c@{}c@{}}
             {\small devotedness(.8259), lovemaking(.8111), heartstrings(.7731), agape(.7582), infatuation(.7415),}\\
             {\small cherish(.7072), eff(.7072), dearest(.6956), do\_it(.6905), fornicate(.6843)}
        \end{tabular} \\
        \cline{2-3}
                            & {\small +SERetro$_6$} &
        \begin{tabular}{@{}c@{}c@{}}
             {\small devotedness(.8229), lovemaking(.8132), heartstrings(.7775), agape(.7627), infatuation(.7499),}\\
             {\small cherish(.7194), eff(.7130), dearest(.7039), do\_it(.6950), fornicate(.6827)}
        \end{tabular} \\
        \cline{2-3}
    \hline
    \multirow{7}{*}{{\small\tt soo}}    & Raw &
        \begin{tabular}{@{}c@{}c@{}}
             {\small sooo(.8394), soooo(.7938), sooooo(.7715), soooooo(.7359), sooooooo(.6844),}\\
             {\small haha(.6574), hahah(.6320), damn(.6247), omg(.6244), hahaha(.6219)}
        \end{tabular} \\
        \cline{2-3}
                            & {+ Retro} &
        \begin{tabular}{@{}c@{}c@{}}
             {\small sooo(.8394), soooo(.7938), sooooo(.7715), soooooo(.7359), sooooooo(.6844), soooooooo(.6896)}\\
             {\small haha(.6574), hahah(.6320), omg(.6244), hahaha(.6219), sooooooooo(.6189)}
        \end{tabular} \\
        \cline{2-3}
                            & {\small +SRExtro$_6$} &
        \begin{tabular}{@{}c@{}c@{}}
             {\small sooo(.7992), soooo(.7701), sooooo(.7570), soooooo(.7339), sooooooo(.7159), sooooooooo(.6838)}\\
             {\small  soooooooo(.6602),  sooooooooooo(.6469), soooooooooo(.6341), tooo(.6293)}
        \end{tabular} \\
        \cline{2-3}
                            & {\small +SERetro$_6$} &
        \begin{tabular}{@{}c@{}c@{}}
             {\small sooo(.8061), soooo(.7559), sooooo(.7413), soooooo(.7167), sooooooo(.6920), sooooooooo(.6521)}\\
             {\small  soooooooo(.6334), sooooooooooo(.6127), tooo(.6089), soooooooooo(.6081)}
        \end{tabular} \\
    \hlineB{3}
    \end{tabular}
    \caption{List of top-10 nearest words of cue words in different post-processing methods. We show cosine similarity scores of two words included in semantic lexicon ({\tt love}) or not ({\tt soo}).}\label{tab:3}
    \end{table*}
    
    \begin{table*}[ht!] \centering
    \begin{tabular}{|l||C{0.9cm}|C{0.9cm}||C{0.9cm}||C{0.9cm}|C{0.9cm}||C{0.9cm}|}
    \hlineB{3}
        Model & {\small \bf MEN} & {\small \bf WS} & {\small \bf Gen.} & {\small \bf SL} & {\small \bf SV} & {\small \bf Spec.} \\
    \hlineB{3}
    \tt\small Retrofitting (Syn) & .7305 & .5332 & .6319 & .4644 & .3017 & .3831 \\
    \hline
    \tt\small Counter-fitting (Syn) & .7149 & .5075 & .6112 & .4143 & .2845 & .3494 \\
    \hline
    \tt\small Counter-fitting (Syn+Ant) & .6898 & .4633 & .5766 & .5415  & .4167 & .4791 \\
    \hline
    \tt\small ATTRACT-REPEL (Syn) & .7156 & .5921 & .6539 & .5672 & .4416 & .5044 \\
    \hline
    \tt\small ATTRACT-REPEL (Syn+Ant) & .7013 & .5523 & .6268 & \bf .6397 & .5463 & .5930 \\
    \hline
    \tt\small ER-CNT$^{*}$ (Syn) & - & - & - & .465 & .339 & .402 \\
    \hline
    \tt\small ER-CNT$^{*}$ (Syn+Ant) & - & - & - & .582 & .439 & .5105 \\
    \hlineB{3}
    \tt\small Unsupervised Extro$_3$ () & \bf .8320 & .6734 & \bf .7527 & .4844 & .3501 & .4173 \\
    \hline
    \tt\small Extro$_6$ (Syn) & .8238 & \bf .6799 & \bf .7519 & .4990 & .3696 & .4343 \\
    \hline
    \tt\small Stepwise RExtro$_6$ (Syn) & .6724 & .5359 & .6042 & .6119 & .5950 & .6035 \\
    \hline
    \tt\small Stepwise ERetro$_6$ (Syn) & .6942 & .5266 & .6104 & \bf .6195 & .5995 & \bf .6095 \\
    \hline
    % \tt\small Stepwise ERetro$_8$ (Syn) & .6697 & .5275 & .5986 & .6055 & \bf .6028 & .6042 \\
    % \hline
    \tt\small Unsupervised + SERetro$_7$ (Syn) & .6834 & .5399 & .6117 & .6169 & \bf .6020 &  \bf .6095 \\
    \hlineB{3}
    \end{tabular}
    \caption{Comparison of our methods with other retrofitting models. We combine train (or dev) set and test set of word similarity dataset together solely for evaluation. We use GloVe with synonym pairs ({\tt Syn}) in WordNet$_{all}$ lexicon and their antonym pair ({\tt Ant}) if the model uses antonyms as well. () means that no external resources are used. The github codes of $^{*}$ER-CNT are under-development so we report the results from their papers.}\label{tab:4}
    \end{table*}
    
\section{Results}
    We compare our best results with previous retrofitting models in Table~\ref{tab:4}. We define the average similarity score of SimLex-999 (SL) and SimVerb-3500 (SV) as specialization score (Spec), in which previous works~\cite{mrkvsic2017semantic,glavavs2018explicit} have tried to increase the performance. The average score of the MEN-3k (MEN) and WordSim-353 are defined as generalization score (Gen) because MEN and WS include words that are not a part of WordNet$_{all}$ lexicon.
    Our methods, Stepwise RExtro and Stepwise ERetro, significantly outperform state-of-the-art retrofitting models despite using only synonyms.\\
    % Furthermore, if we combine extrofitting with retrofitting in greedy way, the result could be further improved.\\
    Although ATTRACT-REPEL~\cite{mrkvsic2017semantic} is better than our methods on SimLex-999, we specialize the word vector with only synonyms, thus using less external resources than ATTRACT-REPEL. Second, \citeauthor{glavavs2018explicit} showed that ATTRACT-REPEL specializes only words seen in semantic lexicons, whereas our methods include the strong point of ER-CNT~\cite{glavavs2018explicit} that is able to enrich word vectors not included in semantic lexicons, by making non-synonyms distant to each other. Lastly, ATTRACT-REPEL cannot use GloVe without preprocessing, because of the limitation of memory allocation. This constraint is critical when we use the pretrained word vectors to the conventional tasks which have large amount of vocabularies.
\section{Downstream Task}
    We utilize our results on text classification tasks, which is the basis of conventional NLP tasks.
    
\subsection{Datasets}
    We use 2 topic classification datasets; {\bf DBpedia ontology}~\cite{lehmann2015dbpedia}, {\bf Yahoo!Answers}~\cite{chang2008importance}, and 1 sentiment classification dataset; {\bf Yelp reviews}. We utilize Yahoo!Answer dataset for 2 different tasks, classifying upper-level categories and classifying lower-level categories, respectively.
    
    \begin{table*}[ht!] \centering
    \begin{tabular}{|l||C{1.8cm}|C{1.8cm}|C{1.8cm}|C{1.8cm}|}
    \hlineB{3}
        \multicolumn{5}{l}{\tt NOT Trainable Word Vectors} \\
        \hlineB{3}
        & \bf DBpedia & \bf \begin{tabular}{@{}c@{}} {\small YahooAnswer} \\ {\small (Upper) } \end{tabular} & \bf \begin{tabular}{@{}c@{}} {\small YahooAnswer} \\ {\small (Lower) } \end{tabular} & \bf Yelp review\\
    \hlineB{3}
    % {\small\tt (1) Without Pretrained} & 0.9740 & 0.6282 & 0.4095 & 0.6719 \\
    % {\small\tt (1) GloVe} & .8544 & .4692 & .3154 & .6528 \\
    {\small\tt (1) GloVe} & .9817 & .7296 & .4990 & .6712 \\
    \hline
    % {\small\tt (2) Retrofit(GloVe)} & .8660 & .4750 & .2954 & .6501 \\
    {\small\tt (2) Retrofit(GloVe)} & .8394 & .4647 & .2941 & .6397 \\
    \hline
    % {\small\tt (3) Counter-fit(GloVe)} & .7713 & .3794 & .2118 & .6156 \\
    {\small\tt (3) Counter-fit(GloVe)} & .7821 & .3807 & .2034 & .6164 \\
    \hline
    % {\small\tt (4) Extro$_1$(GloVe)} & \bf .9864 & .7349 & .5255 & \bf .6804 \\
    {\small\tt (4) Extro$_1$(GloVe)} & \bf .9866 & \bf .7389 & \bf .5253 & .6798 \\
    \hline
    % {\small\tt (5) Unsupervised Extro$_2$(GloVe)} & \bf .9864 & .7344 & \bf .5265 & \bf .6801 \\
    {\small\tt (5) Unsupervised Extro$_3$(GloVe)} & .9860 & .7371 & .5222 & .6804 \\
    \hline
    {\small\tt (6) Extro$_6$(GloVe)} & .9863 & .7370 & .5218 & \bf .6809 \\
    \hline
    % {\small\tt (5) RExtro$_{11}$(GloVe)} & \bf 0.9862 & 0.7347 & \bf 0.5204 & \bf 0.6807 \\
    % \hline
    % {\small\tt (6) ERetro$_{22}$(GloVe)} & 0.9852 & 0.7267 & 0.5140 & 0.6793 \\
    % \hline
    {\small\tt (7) Stepwise RExtro$_{6}$(GloVe)} & .9846 & .7084 & .4894 & .6773 \\
    \hline
    {\small\tt (8) Stepwise ERetro$_{6}$(GloVe)} & .9846 & .7065 & .4950 & .6777 \\
    \hlineB{3}
    \end{tabular}
    
    \begin{tabular}{|l||C{1.8cm}|C{1.8cm}|C{1.8cm}|C{1.8cm}|}
    \hlineB{3}
        \multicolumn{5}{l}{\tt Trainable Word Vectors} \\
        \hlineB{3}
        & \bf DBpedia & \bf \begin{tabular}{@{}c@{}} {\small YahooAnswer} \\ {\small (Upper) } \end{tabular} & \bf \begin{tabular}{@{}c@{}} {\small YahooAnswer} \\ {\small (Lower) } \end{tabular} & \bf Yelp review\\
    \hlineB{3}
    % {\small\tt (1) Without Pretrained} & 0.9822 & 0.6687 & 0.4392 & 0.6796 \\
    \hline
    % {\small\tt (1) GloVe} & .9861 & .6960 & .4592 & .6805 \\
    % \hline
    {\small\tt (1) GloVe} & .9870 & .7327 & .5173 & .6798 \\
    \hline
    % {\small\tt (2) Retrofit(GloVe)} & .9785 & .6689 & .4609 & .6780 \\
    {\small\tt (2) Retrofit(GloVe)} & .9770 & .6473 & .4027 & .6797 \\
    \hline
    % {\small\tt (3) Counter-fit(GloVe)} & .9798 & .6555 & .4339 & .6799 \\
    {\small\tt (3) Counter-fit(GloVe)} & .9821 & .6381 & .4108 & .6778 \\
    \hline
    % {\small\tt (4) Extro$_1$(GloVe)} & \bf .9874 & .7419 & \bf .5319 & .6845 \\
    {\small\tt (4) Extro$_1$(GloVe)} & \bf .9875 & .7493 & .5288 & .6836 \\
    \hline
    % {\small\tt (5) Unsupervised Extro$_2$(GloVe)} & \bf .9869 & \bf .7425 & \bf .5309 & .6842 \\
    {\small\tt (5) Unsupervised Extro$_3$(GloVe)} & \bf .9875 & \bf .7499 & .5284 & .6827 \\
    \hline
    {\small\tt (6) Extro$_6$(GloVe)} & .9873 & .7473 & \bf .5303 & \bf .6844 \\
    \hline
    % {\small\tt (5) RExtro$_{11}$(GloVe)} & \bf 0.9876 & 0.7423 & 0.5255 & 0.6859 \\
    % \hline
    % {\small\tt (6) ERetro$_{22}$(GloVe)} & \bf 0.9875 & 0.7328 & 0.5205 & 0.6862 \\
    % \hline
    {\small\tt (7) Stepwise RExtro$_{6}$(GloVe)} & .9859 & .7201 & .4995 & .6836 \\
    \hline
    {\small\tt (8) Stepwise ERetro$_{6}$(GloVe)} & .9857 & .7234 & .4996 & .6834 \\
    \hlineB{3}
    \end{tabular}
    \caption{10 times average accuracy of TextCNN classifiers initialized by differently post-processed word vector.}\label{tab:5}
    \end{table*}
    
\subsection{Classifier}
    Since we believe that keeping the sequence of words is important, we build simple TextCNN~\cite{kim2014convolutional} rather than building a classifier based on Bag-of-Words (BoW) as ~\citeauthor{faruqui2015retrofitting} did, since BoW neglects the word sequences by averaging all the word vectors.\\
    We use the first 100 words as input sequences, and the classifier consists of 2 convolutional layers with the channel size of 32 and 16, respectively. We adopt the multi-channel approach, implementing 4 different sizes of kernels--2, 3, 4, and 5. We concatenate them after every max-pooling layer. The learned kernels go through an activation function, ReLU~\cite{hahnloser2000digital}, and are max-pooled. We set the size of word embedding to 300, optimizer to Adam~\cite{kingma2014adam} with learning rate 0.001, using early-stopping.
    
\subsection{Experiment}
    We experiment with our methods in 2 different settings: fixed word vectors, or trainable word vectors. When the word vectors are fixed, we can evaluate the usefulness of the word vectors per se. With the trainable word vectors, we can see the improvement of the classification performance when initialized with the enriched word vectors.
    
\subsection{Results}
    In each setting, we report the performance of the classifier in Table~\ref{tab:5}.
    % (1) without any pretrained word vector, (2) with GloVe, (3) GloVe with retrofitting, (4) GloVe with counter-fitting, (5) GloVe with extrofitting, (6) GloVe with stacked extrofitting, (7) GloVe with Stepwise RExtro, and (8) GloVe with Stepwise ERetro. The results are presented in Table~\ref{tab:10}.
    classification results with the generalized word vectors, (4), (5) and (6), are better than the results with the specialized word vectors, (2), (3), (7) and (8) both when the word vectors are fixed and trainable. Even the specialized word vectors degrade the classification performance when compared to original GloVe. The classifier initialized with (5) unsupervised extrofitting outperforms the original GloVe.\\ 
    The performance gap between (4) simple extrofitting and the enriched word vectors, (5) and (6), is small but the result that the classifier with (5) unsupervised extrofitting performs on par with (4) extrofitting is noticeable. Also, our specialized word vectors, (7) and (8), degrade less performance than (2) and (3).\\
    Consequently, we claim that our methods are word vector enrichment, which makes the performance gain on conventional NLP tasks, i.e., word vector post-processing method for general purpose.\\
    Other word vector specialization models, ATTRACT-REPEL and ER-CNT, cannot be compared with our methods since the models cannot use GloVe without preprocessing and the github codes are under-development, respectively. In addition, our methods focus on retrofitting with semantic information, so it is unfair to compare our methods with contextual representations such as ELMo~\cite{peters2018deep}.
    % The results imply that our methods successfully inject semantic information into the word vectors but the gap could be filled by learning and updating the word vectors.
    % in sentiment classification (Yelp review). This might be from that WordNet$_{all}$ contains numerous emotion words. The result implies that although generalized word vectors perform better in general, specialized word vectors can be useful for domain-specific tasks if we have enough specialized semantic lexicons.

\section{Conclusion}
    We develop retrofitting models that one is able to enrich word vector without semantic lexicon ({\em unsupervised extrofitting}) and the other is using in-depth expansional retrofitting ({\em deep extrofitting}). We show that unsupervised extrofitting improves the performance on overall word similarity tasks compared to GloVe and present its application as word vector ensemble. Next, we show that in-depth combinations of extrofitting with retrofitting outperform previous state-of-the-art models in word vector specialization.
    % , specializing on SimLex-999 and SimVerb-3500 with only synonyms and generalizing on MEN-3k and WordSim-353.\\
    % Also, we can see not only that extrofitting helps retrofitting find new vector space on specialization that prevents retrofitting from converging in a few iterations but also retrofitting helps extrofitting to strongly collect word vectors. Our method is dependent on the distribution of pretrained word vectors and synonym pairs, and does not need antonym pairs, hyperparameters, and explicit mapping functions. As a future work, we will further research our method to utilize antonym pairs as well.
    
% \section*{Acknowledgments}

% The acknowledgments should go immediately before the references.  Do
% not number the acknowledgments section. Do not include this section
% when submitting your paper for review. \\

\bibliography{naaclhlt2019}

\begin{thebibliography}{30}
\expandafter\ifx\csname natexlab\endcsname\relax\def\natexlab#1{#1}\fi

\bibitem[{Baker et~al.(1998)Baker, Fillmore, and Lowe}]{baker1998berkeley}
Collin~F Baker, Charles~J Fillmore, and John~B Lowe. 1998.
\newblock The berkeley framenet project.
\newblock In \emph{Proceedings of the 17th international conference on
  Computational linguistics-Volume 1}, pages 86--90. Association for
  Computational Linguistics.

\bibitem[{Bojanowski et~al.(2016)Bojanowski, Grave, Joulin, and
  Mikolov}]{bojanowski2016enriching}
Piotr Bojanowski, Edouard Grave, Armand Joulin, and Tomas Mikolov. 2016.
\newblock Enriching word vectors with subword information.
\newblock \emph{arXiv preprint arXiv:1607.04606}.

\bibitem[{Bruni et~al.(2014)Bruni, Tram, Baroni et~al.}]{bruni2014multimodal}
Elia Bruni, N~Tram, Marco Baroni, et~al. 2014.
\newblock Multimodal distributional semantics.
\newblock \emph{The Journal of Artificial Intelligence Research}, 49:1--47.

\bibitem[{Camacho-Collados et~al.(2017)Camacho-Collados, Pilehvar, Collier, and
  Navigli}]{camacho2017semeval}
Jose Camacho-Collados, Mohammad~Taher Pilehvar, Nigel Collier, and Roberto
  Navigli. 2017.
\newblock Semeval-2017 task 2: Multilingual and cross-lingual semantic word
  similarity.
\newblock In \emph{Proceedings of the 11th International Workshop on Semantic
  Evaluation (SemEval-2017)}, pages 15--26.

\bibitem[{Cer et~al.(2017)Cer, Diab, Agirre, Lopez-Gazpio, and
  Specia}]{cer2017semeval}
Daniel Cer, Mona Diab, Eneko Agirre, Inigo Lopez-Gazpio, and Lucia Specia.
  2017.
\newblock Semeval-2017 task 1: Semantic textual similarity-multilingual and
  cross-lingual focused evaluation.
\newblock \emph{arXiv preprint arXiv:1708.00055}.

\bibitem[{Chang et~al.(2008)Chang, Ratinov, Roth, and
  Srikumar}]{chang2008importance}
Ming-Wei Chang, Lev-Arie Ratinov, Dan Roth, and Vivek Srikumar. 2008.
\newblock Importance of semantic representation: Dataless classification.
\newblock In \emph{AAAI}, volume~2, pages 830--835.

\bibitem[{Daniel(1990)}]{daniel1990spearman}
Wayne~W Daniel. 1990.
\newblock Spearman rank correlation coefficient.
\newblock \emph{Applied nonparametric statistics}, pages 358--365.

\bibitem[{Faruqui et~al.(2015)Faruqui, Dodge, Jauhar, Dyer, Hovy, and
  Smith}]{faruqui2015retrofitting}
Manaal Faruqui, Jesse Dodge, Sujay~Kumar Jauhar, Chris Dyer, Eduard Hovy, and
  Noah~A Smith. 2015.
\newblock Retrofitting word vectors to semantic lexicons.
\newblock In \emph{Proceedings of the 2015 Conference of the North American
  Chapter of the Association for Computational Linguistics: Human Language
  Technologies}, pages 1606--1615.

\bibitem[{Finkelstein et~al.(2001)Finkelstein, Gabrilovich, Matias, Rivlin,
  Solan, Wolfman, and Ruppin}]{finkelstein2001placing}
Lev Finkelstein, Evgeniy Gabrilovich, Yossi Matias, Ehud Rivlin, Zach Solan,
  Gadi Wolfman, and Eytan Ruppin. 2001.
\newblock Placing search in context: The concept revisited.
\newblock In \emph{Proceedings of the 10th international conference on World
  Wide Web}, pages 406--414. ACM.

\bibitem[{Ganitkevitch et~al.(2013)Ganitkevitch, Van~Durme, and
  Callison-Burch}]{ganitkevitch2013ppdb}
Juri Ganitkevitch, Benjamin Van~Durme, and Chris Callison-Burch. 2013.
\newblock Ppdb: The paraphrase database.
\newblock In \emph{Proceedings of the 2013 Conference of the North American
  Chapter of the Association for Computational Linguistics: Human Language
  Technologies}, pages 758--764.

\bibitem[{Gerz et~al.(2016)Gerz, Vuli{\'c}, Hill, Reichart, and
  Korhonen}]{gerz2016simverb}
Daniela Gerz, Ivan Vuli{\'c}, Felix Hill, Roi Reichart, and Anna Korhonen.
  2016.
\newblock Simverb-3500: A large-scale evaluation set of verb similarity.
\newblock In \emph{Proceedings of the 2016 Conference on Empirical Methods in
  Natural Language Processing}, pages 2173--2182.

\bibitem[{Glava{\v{s}} and Vuli{\'c}(2018)}]{glavavs2018explicit}
Goran Glava{\v{s}} and Ivan Vuli{\'c}. 2018.
\newblock Explicit retrofitting of distributional word vectors.
\newblock In \emph{Proceedings of the 56th Annual Meeting of the Association
  for Computational Linguistics (Volume 1: Long Papers)}, volume~1, pages
  34--45.

\bibitem[{Hahnloser et~al.(2000)Hahnloser, Sarpeshkar, Mahowald, Douglas, and
  Seung}]{hahnloser2000digital}
Richard~HR Hahnloser, Rahul Sarpeshkar, Misha~A Mahowald, Rodney~J Douglas, and
  H~Sebastian Seung. 2000.
\newblock Digital selection and analogue amplification coexist in a
  cortex-inspired silicon circuit.
\newblock \emph{Nature}, 405(6789):947.

\bibitem[{Hill et~al.(2015)Hill, Reichart, and Korhonen}]{hill2015simlex}
Felix Hill, Roi Reichart, and Anna Korhonen. 2015.
\newblock Simlex-999: Evaluating semantic models with (genuine) similarity
  estimation.
\newblock \emph{Computational Linguistics}, 41(4):665--695.

\bibitem[{Jo and Choi(2018)}]{jo2018extrofitting}
Hwiyeol Jo and Stanley~Jungkyu Choi. 2018.
\newblock Extrofitting: Enriching word representation and its vector space with
  semantic lexicons.
\newblock \emph{arXiv preprint arXiv:1804.07946}.

\bibitem[{Kiela et~al.(2015)Kiela, Hill, and Clark}]{kiela2015specializing}
Douwe Kiela, Felix Hill, and Stephen Clark. 2015.
\newblock Specializing word embeddings for similarity or relatedness.
\newblock In \emph{Proceedings of the 2015 Conference on Empirical Methods in
  Natural Language Processing}, pages 2044--2048.

\bibitem[{Kim(2014)}]{kim2014convolutional}
Yoon Kim. 2014.
\newblock Convolutional neural networks for sentence classification.
\newblock In \emph{Proceedings of the 2014 Conference on Empirical Methods in
  Natural Language Processing (EMNLP)}, pages 1746--1751.

\bibitem[{Kingma and Ba(2014)}]{kingma2014adam}
Diederik~P Kingma and Jimmy Ba. 2014.
\newblock Adam: A method for stochastic optimization.
\newblock \emph{arXiv preprint arXiv:1412.6980}.

\bibitem[{Landauer and Dumais(1997)}]{landauer1997solution}
Thomas~K Landauer and Susan~T Dumais. 1997.
\newblock A solution to plato's problem: The latent semantic analysis theory of
  acquisition, induction, and representation of knowledge.
\newblock \emph{Psychological review}, 104(2):211.

\bibitem[{Lehmann et~al.(2015)Lehmann, Isele, Jakob, Jentzsch, Kontokostas,
  Mendes, Hellmann, Morsey, Van~Kleef, Auer et~al.}]{lehmann2015dbpedia}
Jens Lehmann, Robert Isele, Max Jakob, Anja Jentzsch, Dimitris Kontokostas,
  Pablo~N Mendes, Sebastian Hellmann, Mohamed Morsey, Patrick Van~Kleef,
  S{\"o}ren Auer, et~al. 2015.
\newblock Dbpedia--a large-scale, multilingual knowledge base extracted from
  wikipedia.
\newblock \emph{Semantic Web}, 6(2):167--195.

\bibitem[{Lenci(2018)}]{lenci2018distributional}
Alessandro Lenci. 2018.
\newblock Distributional models of word meaning.
\newblock \emph{Annual review of Linguistics}, 4:151--171.

\bibitem[{Maaten and Hinton(2008)}]{maaten2008visualizing}
Laurens van~der Maaten and Geoffrey Hinton. 2008.
\newblock Visualizing data using t-sne.
\newblock \emph{Journal of machine learning research}, 9(Nov):2579--2605.

\bibitem[{Mikolov et~al.(2013{\natexlab{a}})Mikolov, Chen, Corrado, and
  Dean}]{mikolov2013efficient}
Tomas Mikolov, Kai Chen, Greg Corrado, and Jeffrey Dean. 2013{\natexlab{a}}.
\newblock Efficient estimation of word representations in vector space.
\newblock \emph{arXiv preprint arXiv:1301.3781}.

\bibitem[{Mikolov et~al.(2013{\natexlab{b}})Mikolov, Sutskever, Chen, Corrado,
  and Dean}]{mikolov2013distributed}
Tomas Mikolov, Ilya Sutskever, Kai Chen, Greg~S Corrado, and Jeff Dean.
  2013{\natexlab{b}}.
\newblock Distributed representations of words and phrases and their
  compositionality.
\newblock In \emph{Advances in neural information processing systems}, pages
  3111--3119.

\bibitem[{Miller(1995)}]{miller1995wordnet}
George~A Miller. 1995.
\newblock Wordnet: a lexical database for english.
\newblock \emph{Communications of the ACM}, 38(11):39--41.

\bibitem[{Mrk{\v{s}}i{\'c} et~al.(2016)Mrk{\v{s}}i{\'c}, S{\'e}aghdha, Thomson,
  Ga{\v{s}}i{\'c}, Rojas-Barahona, Su, Vandyke, Wen, and
  Young}]{mrkvsic2016counter}
Nikola Mrk{\v{s}}i{\'c}, Diarmuid~O S{\'e}aghdha, Blaise Thomson, Milica
  Ga{\v{s}}i{\'c}, Lina Rojas-Barahona, Pei-Hao Su, David Vandyke, Tsung-Hsien
  Wen, and Steve Young. 2016.
\newblock Counter-fitting word vectors to linguistic constraints.
\newblock \emph{arXiv preprint arXiv:1603.00892}.

\bibitem[{Mrk{\v{s}}i{\'c} et~al.(2017)Mrk{\v{s}}i{\'c}, Vuli{\'c},
  S{\'e}aghdha, Leviant, Reichart, Ga{\v{s}}i{\'c}, Korhonen, and
  Young}]{mrkvsic2017semantic}
Nikola Mrk{\v{s}}i{\'c}, Ivan Vuli{\'c}, Diarmuid~{\'O} S{\'e}aghdha, Ira
  Leviant, Roi Reichart, Milica Ga{\v{s}}i{\'c}, Anna Korhonen, and Steve
  Young. 2017.
\newblock Semantic specialization of distributional word vector spaces using
  monolingual and cross-lingual constraints.
\newblock \emph{Transactions of the Association for Computational Linguistics},
  5:309--324.

\bibitem[{Pennington et~al.(2014)Pennington, Socher, and
  Manning}]{pennington2014glove}
Jeffrey Pennington, Richard Socher, and Christopher Manning. 2014.
\newblock Glove: Global vectors for word representation.
\newblock In \emph{Proceedings of the 2014 conference on empirical methods in
  natural language processing (EMNLP)}, pages 1532--1543.

\bibitem[{Peters et~al.(2018)Peters, Neumann, Iyyer, Gardner, Clark, Lee, and
  Zettlemoyer}]{peters2018deep}
Matthew Peters, Mark Neumann, Mohit Iyyer, Matt Gardner, Christopher Clark,
  Kenton Lee, and Luke Zettlemoyer. 2018.
\newblock Deep contextualized word representations.
\newblock In \emph{Proceedings of the 2018 Conference of the North American
  Chapter of the Association for Computational Linguistics: Human Language
  Technologies, Volume 1 (Long Papers)}, volume~1, pages 2227--2237.

\bibitem[{Wieting et~al.(2015)Wieting, Bansal, Gimpel, Livescu, and
  Roth}]{wieting2015paraphrase}
John Wieting, Mohit Bansal, Kevin Gimpel, Karen Livescu, and Dan Roth. 2015.
\newblock From paraphrase database to compositional paraphrase model and back.
\newblock \emph{Transactions of the Association for Computational Linguistics},
  3:345--358.

\end{thebibliography}
\bibliographystyle{acl_natbib}

\clearpage

\appendix
\renewcommand\thefigure{\thesection.\arabic{figure}}
\renewcommand\thetable{\thesection.\arabic{table}}
\setcounter{figure}{0}
\setcounter{table}{0}
\section{Appendices}
\label{sec:appendix}
\subsection{Retrofitting with extracted semantic information}\label{appendix:a}

\begin{minipage}{\textwidth}
    \centering
    \begin{tabular}{|l||c|c|c|c|} \hlineB{3}
    & \bf MEN & \bf WS & \bf SL & \bf SV \\ \hlineB{3}
    GloVe & .7435 & .5516 & .3738 & .2264 \\ \hline
    Retro with GloVe(100) & .7435 & .5521 & .3738 & .2245 \\ \hline
    Extro with GloVe(100) & .8271 & .6506 & .4754 & .3382 \\
    \hlineB{3}
    \end{tabular}
    \captionof{table}{Spearman's correlation of post-processed GloVe with semantic information extracted from GloVe in latent space. GloVe(N) means extracted semantic information from GloVe in N-dimensional latent space.}
    \label{tab:a}
\end{minipage}

\subsection{List of top-10 nearest words of cue words}\label{appendix:b}

\begin{minipage}{\textwidth}
    % \begin{table*}[ht!] \centering
    \begin{tabular}{|c||c|c|}
    \hlineB{3}
        {\small Word} & {\small Method} & Top-10 Nearest Words(Cosine Similarity Score) \\
    \hlineB{3}
    \multirow{9}{*}{{\tt love}}   & Raw &
        \begin{tabular}{@{}c@{}c@{}}
             {\small loved(.7745), i(.7338), loves(.7311), know(.7286), loving(.7263),}\\
             {\small really(.7196), always(.7193), want(.7192), hope(.7127), think(.7110)}
        \end{tabular} \\
        \cline{2-3}
                            & {+ Extro$_1$} &
        \begin{tabular}{@{}c@{}c@{}}
             {\small adore(.5958), hate(.5925), loved(.5786), luv(.5406), loooove(.5290),}\\
             {\small looooove(.5217), loveeee(.5177), want(.5166), loving(.5157), looove(.5071)}
        \end{tabular} \\
        \cline{2-3}
                            & {+ Extro$_2$} &
        \begin{tabular}{@{}c@{}c@{}}
             {\small adore(.5798), hate(.5738), loved(.5572), luv(.5287), loooove(.5254),}\\
             {\small looooove(.5227), loveeee(.5210), looove(.5042), loooooove(.4987), loving(.4948)}
        \end{tabular} \\
        \cline{2-3}
                            & {+ Extro$_6$} &
        \begin{tabular}{@{}c@{}c@{}}
             {\small adore(.5841), hate(.5636), loved(.5518), luv(.5285), loooove(.5282),}\\
             {\small looooove(.5266), loveeee(.5251), looove(.5072), loooooove(.5029), loadsss(.4967)}
        \end{tabular} \\
        \cline{2-3}
                            & \begin{tabular}{@{}c@{}c@{}} {\small + Unsup.}\\{\small Extro$_3$} \end{tabular} &
        \begin{tabular}{@{}c@{}c@{}}
             {\small loves(.6030), loving(.5875), loved(.5805), luv(.5339), adore(.5296),}\\
             {\small friendship(.5284), passion(.5249), likes(.4918), loadsss(.4896), affection(.4833)}
        \end{tabular} \\
        % \cline{2-3}
                            % & \begin{tabular}{@{}c@{}c@{}} {\small + Unsup.}\\{\small ERetro$_7$} \end{tabular} &
        % \begin{tabular}{@{}c@{}c@{}}
            %  {\small devotedness(.8310), lovemaking(.8150), agape(.8061), heartstrings(.7976), infatuation(.7568),}\\
            %  {\small eff(.7434), do\_it(.7341), gaping(.7331), cherish(.7197), fornicate(.7193)}
        % \end{tabular} \\
    \hline
    \multirow{9}{*}{{\tt soo}}    & Raw &
        \begin{tabular}{@{}c@{}c@{}}
             {\small sooo(.8394), soooo(.7938), sooooo(.7715), soooooo(.7359), sooooooo(.6844),}\\
             {\small haha(.6574), hahah(.6320), damn(.6247), omg(.6244), hahaha(.6219)}
        \end{tabular} \\
        \cline{2-3}
                            & {+ Extro$_1$} &
        \begin{tabular}{@{}c@{}c@{}}
             {\small sooo(.8307), soooo(.7870), sooooo(.7754), soooooo(.7554), sooooooo(.7260), soooooooo(.6884),}\\
             {\small  sooooooooo(.6818), soooooooooo(.6545), tooo(.6502), sooooooooooo(.6453)}
        \end{tabular} \\
        \cline{2-3}
                            & {+ Extro$_2$} &
        \begin{tabular}{@{}c@{}c@{}}
             {\small sooo(.8284), soooo(.7843), sooooo(.7723), soooooo(.7525), sooooooo(.7221), soooooooo(.6849),}\\
             {\small  sooooooooo(.6790), soooooooooo(.6516), tooo(.6445), sooooooooooo(.6424)}
        \end{tabular} \\
        \cline{2-3}
                            & {+ Extro$_6$} &
        \begin{tabular}{@{}c@{}c@{}}
             {\small sooo(.8273), soooo(.7828), sooooo(.7707), soooooo(.7508), sooooooo(.7203), soooooooo(.6831),}\\
             {\small  sooooooooo(.6773), soooooooooo(.6498), tooo(.6426), sooooooooooo(.6408)}
        \end{tabular} \\
        \cline{2-3}
                            & \begin{tabular}{@{}c@{}c@{}} {\small + Unsup.}\\{\small Extro$_3$} \end{tabular} &
        \begin{tabular}{@{}c@{}c@{}}
             {\small sooo(.7275), soooo(.6602), sooooo(.6360), soooooo(.5980), sooooooo(.5440), jin(.5328),}\\
             {\small  hyun(.5292), hee(.5136), jung(.5017), soooooooo(.4853)}
        \end{tabular}\\
        % \cline{2-3}
                            % & \begin{tabular}{@{}c@{}c@{}} {\small + Unsup.}\\{\small ERetro$_7$} \end{tabular} &
        % \begin{tabular}{@{}c@{}c@{}}
            %  {\small sooo(.8051), soooo(.7627), sooooo(.7469), soooooo(.7261), sooooooo(.6984), sooooooooo(.6536),}\\
            %  {\small  soooooooo(.6400), soooooooooo(.6186), soooooooooo(.6163), tooo(.6150)}
        % \end{tabular} \\
    \hlineB{3}
    \end{tabular}
    % \caption{}
    % \end{table*}
    \captionof{table}{List of top-10 nearest words of cue words in different post-processing methods. We show cosine similarity scores of two words included in semantic lexicon ({\tt love}) or not ({\tt soo}).}
    \label{tab:b}
\end{minipage}

\end{document}